# UrbanGenAI – Reconstructing Urban Landscapes using Panoptic Segmentation and Diffusion Models

Authors

**Timo Kapsalis**[a] – t.kapsalis@derby.ac.uk

[a]: University of Derby, UK

## Abstract

In contemporary design practices, the integration of computer vision and generative artificial intelligence (genAI) represents a transformative shift towards more interactive and inclusive processes. These technologies offer new dimensions of image analysis and generation, which are particularly relevant in the context of urban landscape reconstruction. This paper presents a novel workflow encapsulated within a prototype application, designed to leverage the synergies between advanced image segmentation and diffusion models for a comprehensive approach to urban design. Our methodology encompasses the OneFormer model for detailed image segmentation and the Stable Diffusion XL (SDXL) diffusion model, implemented through ControlNet, for generating images from textual descriptions. Validation results indicated a high degree of performance by the prototype application, showcasing significant accuracy in both object detection and text-to-image generation. This was evidenced by superior Intersection over Union (IoU) and CLIP scores across iterative evaluations for various categories of urban landscape features. Preliminary testing included utilising UrbanGenAI as an educational tool enhancing the learning experience in design pedagogy, and as a participatory instrument facilitating community-driven urban planning. Early results suggested that UrbanGenAI not only advances the technical frontiers of urban landscape reconstruction but also provides significant pedagogical and participatory planning benefits. The ongoing development of UrbanGenAI aims to further validate its effectiveness across broader contexts and integrate additional features such as real-time feedback mechanisms and 3D modelling capabilities.

<u>Keywords</u>: *generative AI; panoptic image segmentation; diffusion models; urban landscape design; design pedagogy; co-design*

## 1. Introduction

The advent of machine learning and generative artificial intelligence (genAI) has revolutionised numerous fields, with urban design emerging as a particularly fertile ground for these innovative technologies [1,2]. GenAI's ability to analyse complex datasets and generate insightful predictions lends itself well to urban design, where nuanced understanding and foresight are paramount. This convergence of computational prowess and design acumen holds the potential to redefine how urban environments are conceptualised, planned, and realised [3]. It enables designers to navigate the intricate tapestry of urban landscapes with enhanced precision and creativity, leveraging genAI to synthesise vast arrays of environmental, social, and structural data into coherent and sustainable urban visions.

Image segmentation stands as a cornerstone in the realm of computer vision. It is a process that dissects a digital image into constituent segments or pixels, facilitating a more manageable and insightful representation of the image [4]. By assigning pixels to specific labels based on shared attributes, image segmentation transforms the visual complexity of urban landscapes into discrete, analysable components. This segmentation can be instrumental in urban design, where the ability to distinguish between various elements – e.g., buildings, streetscapes, vehicles, and green spaces – is essential. Advanced algorithms within machine learning, especially deep learning, have propelled image segmentation into a sophisticated tool, capable of discerning fine details within urban environments that are critical for spatial analysis, infrastructure planning, and the enhancement of virtual simulations of cities [5,6].

Diffusion models are a subset of genAI models that focus on the task of image synthesis and modification. These models function by first introducing noise into images, effectively distorting or removing original information [7]. Through a deep learning process, the models then develop the capability to reverse this noise addition, gradually restoring or recreating the original pictorial content [7,8]. This method is commonly employed in genAI applications where the goal is to create or edit images. For image editing, or "inpainting", these models work by first introducing noise into specific areas of an image and then iteratively refining these areas by predicting and reversing the noise, effectively "filling in" the missing or damaged parts of the image [9]. The applications of diffusion models are particularly intriguing in the field of urban environment reconstruction. They can be used to visualise potential changes in an urban environment, such as modifying existing buildings, adding new structures, or altering landscapes [10-12]. This capability makes them highly valuable for urban designers and architects who wish to create detailed and realistic visualisations of proposed changes to urban spaces before any real-world changes are made.

The purpose of this paper is to bridge the gap between the theoretical potential and practical application of machine learning in the field of urban design. We introduce a workflow that not only incorporates state-of-the-art computer vision for image segmentation but also leverages the capabilities of diffusion models for urban landscape reconstruction. Based on this framework, a secondary aim is to develop a prototype application, which will process images representing urban landscapes. The application's main function to edit specific sections of these images using only textual descriptions, or "prompts", provided by users. This innovative approach circumvents the need for engaging in advanced, complicated, or specialised image analysis tasks. The essence of this application lies in its ability to leverage the underlying principles of the framework to translate simple text inputs into complex image modifications, thereby simplifying user interaction with design processing tools.

While there have been similar research endeavours to harness the capabilities of text-driven genAI models for urban landscape analysis [13-16], these efforts often exhibit a narrower scope compared to the comprehensive approach presented in our work. For example, previous studies have predominantly concentrated on the reconstruction of buildings alone, offering valuable insights but lacking a holistic view of the urban tapestry [13,15]. Our effort distinguishes itself by not limiting the application to architectural structures but extending it to the intricate interplay of all urban elements. This includes the dynamic flux of vehicles, the verdant sprawl of natural elements, and the ubiquitous presence of street furniture. Encapsulating the urban environment holistically, our approach intends to offer a nuanced tool for detailed reconstructions of urban landscapes. This tool has the potential to be beneficial across various contexts, from educational environments where it can enhance learning, to community co-design initiatives where it can facilitate collaborative urban planning.

## 2. Workflow description

In the proposed workflow (Figure 1), we delineate a multi-step process integrating advanced image segmentation and generative AI techniques. This workflow is designed to modify and reconstruct urban landscape images, thereby facilitating nuanced visualisations in urban design studies.

**1. Image upload by user.** The process initiates with users uploading an urban landscape image to the application. This step is crucial as it provides the foundational data for all subsequent image processing and reconstruction tasks. The significance of the initial image lies in its detail and composition, which directly influence the effectiveness of the segmentation and reconstruction phases.

**2. Panoptic segmentation via OneFormer.** Upon image upload, the OneFormer model undertakes panoptic segmentation [17]. Panoptic segmentation, a comprehensive image analysis technique, differentiates and categorises every pixel in an image into various segments or "masks" [18]. OneFormer, in this context, automatically applies these masks to the urban landscape image, segmenting it into distinct objects and regions [18]. This automatic segmentation is vital for identifying and isolating individual components within the complex urban environment.

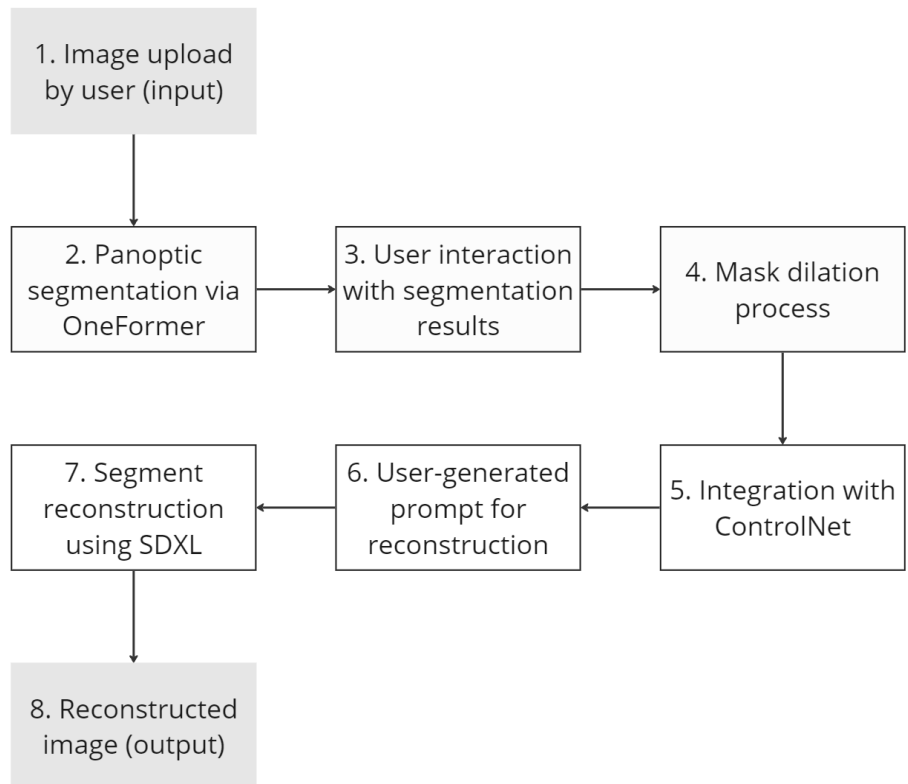

*Figure 1: Schematic representation of the proposed workflow.*

**3. User interaction with segmentation results.** Post-segmentation, the segmented image is presented to the user. Users interact with this output by a selecting specific segment – i.e., an individual object – within the urban landscape. This selection is achieved through a direct click interface, where users identify and choose segments of interest. This step is instrumental in empowering users to focus on particular areas or features within the urban landscape for detailed study or modification.

**4. Mask dilation process.** Following the selection of a segment, a mask dilation process is employed. Mask dilation is a morphological operation that increases the size of the object in the mask, creating a slightly enlarged boundary around the selected segment [19]. The mask of the selected segment is applied and dilated to create a smoother blending with the reconstructed segment. The selected segment, resized and aligned with the dilated mask, is processed using techniques like feathering and colour correction for a natural integration. This dilation is crucial for ensuring a buffer sone that aids in the seamless integration of the reconstructed segment in subsequent steps.

**5. Integration with ControlNet.** The original image, along with the image featuring the dilated mask, is then passed to ControlNet. ControlNet is an advanced framework designed for image reconstruction tasks, including inpainting and generative modifications based on specified conditions [20]. The integration of ControlNet into our workflow is a pivotal step, as it leverages the framework's capabilities to modify and reconstruct urban landscape segments based on the user's input and the segmentation data.

**6. User-generated prompt.** Concurrently, the user provides a textual prompt that describes the desired modifications or reconstructions of the selected segment. This prompt plays a crucial role as it conditions the output of ControlNet, guiding the nature and specifics of the reconstructed image.

**7. Segment reconstruction using SDXL.** ControlNet, utilising the Stable Diffusion XL (SDXL) large-scale diffusion model [21], processes the input image, the dilated mask, and the user's prompt to reconstruct the selected segment. This reconstruction is conditioned both by the spatial information provided by the mask and the descriptive elements of the prompt. The result is a modified segment that aligns with the user's vision while maintaining coherence with the surrounding urban landscape.

**8. Reconstructed image with export and save options for users.** Finally, the reconstructed image is returned to the user, who is then provided with options to export or save the image. This step is essential in the workflow as it allows users to retain and utilise the reconstructed urban landscape for further analysis, presentation, or archival purposes.

## 3. Development environment

In our prototype application, titled *UrbanGenAI*, we have employed a suite of specialised tools to facilitate each step of the workflow. The initial user interaction, involving the upload of an urban landscape image, is managed using a Python-based GUI toolkit, which effectively handles image upload and file I/O operations. For the panoptic segmentation of the uploaded image, we integrate the OneFormer model, utilising TensorFlow to run this advanced segmentation framework efficiently [24].

User interactions with the segmented image, specifically the selection of individual segments, are facilitated through an intuitive desktop interface developed in Python, using the PyQt toolkit. The process of mask dilation, a critical step for ensuring seamless integration in later stages, is carried out using the OpenCV library, which is a staple in image processing tasks [22].

The core of our application's functionality – the image reconstruction process – is powered by the ControlNet framework, and is integrated directly into our desktop application environment. Accompanying this, the user-generated prompts are handled through the same GUI interface, ensuring a cohesive user experience.

The final reconstruction of the image, conditioned by the mask and user prompt, is executed by ControlNet utilising the SDXL model. This sophisticated AI-driven process is central to our application's capability to modify and reimagine urban landscapes. Finally, the application provides options for users to export or save the reconstructed image, a feature implemented using Python's standard file handling libraries, allowing users to retain and utilise their modified landscapes elsewhere.

## 4. Training

We adopted a tailored approach to train the models used in our desktop application for reconstructing urban landscapes. This training process was critical to optimise the models' performance and ensure their relevance and effectiveness in meeting our research objectives.

### 4.1. Training of the OneFormer detection model

For the training of the OneFormer detection model, we employed a specific supervised learning approach known as transfer learning [26]. Transfer learning is a machine learning method where a model developed for a specific task is reused as the starting point for a model on a second task. In our case, the OneFormer model, which had been originally trained on a diverse and extensive dataset, was fine-tuned using the Cityscapes dataset [24]. This dataset, which is rich in annotated urban street scenes, provided the necessary context and variability for the model to adapt and learn specific features relevant to urban landscapes. The transfer learning process involved tweaking the model's parameters to better fit the characteristics of the Cityscapes dataset, enabling the model to effectively recognise and segment various urban elements such as buildings, roads, natural elements, and vehicles.

The advantage of using transfer learning lies in its ability to leverage the knowledge gained by the model in its original training on a large, diverse dataset and apply it to a more specialised task. This not only shortens the training time but also improves the model's accuracy and efficiency in segmenting urban landscape images, making it a suitable choice for our application in urban landscape reconstruction.

### 4.2. Training of the SDXL diffusion model

The SDXL diffusion model, a pivotal component for the reconstruction of selected image segments, was trained using the Dreambooth method [25]. Dreambooth is a personalised training approach that allows for the customisation of generative models with a relatively small dataset. This method is particularly beneficial in fine-tuning generative models to produce results that are closely aligned with specific project requirements or stylistic preferences.

We compiled a dataset consisting of 100 images, each meticulously chosen to encompass a wide array of urban scenarios and features pertinent to urban landscape reconstruction. During the training process, the SDXL model was subjected to 14 learning epochs. An epoch in machine learning is a term used to describe one cycle through the full training dataset. By iterating through these 14 epochs, the SDXL model progressively learned and adapted to the specific styles, textures, and architectural elements that are characteristic of urban landscapes. This repetitive learning process was vital for the model to deeply internalise the nuances of the dataset, thereby enhancing its ability to generate realistic and contextually coherent urban landscape reconstructions.

The selection of 14 epochs was a strategic decision, balancing the need for sufficient training to achieve high fidelity in the generated images, while avoiding overfitting where the model becomes too narrowly tuned to the training dataset. Overfitting can lead to a loss of generality and a decrease in the model's ability to adapt to new, unseen images. Thus, the chosen number of epochs ensured that the SDXL model was effectively trained to our project's specific requirements without compromising its versatility and adaptability in practical applications.

### 4.3. Computational setup

The computational setup for training the OneFormer and SDXL model is presented in Table 1 below:

| **CPU** | Intel Core i9-12900 3.8 GHs, 16-core processor |
|---|---|
| **GPU** | NVIDIA GeForce RTX 3080, 12 GB VRAM |
| **RAM** | 64 GB DDR5 memory |
| **Storage** | 1 TB SSD storage |
| **Operating system** | Windows 11 |

*Table 1: Characteristics of computational setup*

## 5. Validation

In the realm of machine learning, validation is a cornerstone for establishing the efficacy and reliability of developed models. It is a crucial phase underscores the accuracy of the models and instils confidence in their practical deployment. In our study, we have conducted a rigorous two-fold validation process. The first type of validation focuses on the precision of the panoptic segmentation model, i.e., OneFormer, in accurately detecting individual objects within complex urban landscapes. The second strand of validation assesses the fidelity of the fine-tuned image generation model, i.e., SDXL, in creating images from textual descriptions. Together, these validation approaches are instrumental in quantifying the performance of our models, ensuring that the synthesised urban landscapes are visually compelling and contextually coherent with the textual inputs provided.

### 5.1. Accuracy of individual object detection

The validation of individual object detection accuracy is a critical aspect of our research, primarily to ensure the reliability and precision of the OneFormer detection model in segmenting urban landscape images. Accurate object detection is foundational to our project, as it directly influences the quality of the subsequent image generation process. The segmented objects serve as the basis for further processing and reconstruction, thus necessitating high accuracy in segmentation to achieve coherent and realistic modifications of the urban landscape.

For validating the accuracy of individual object detection, we have chosen the Intersection over Union (IoU) metric. IoU is a widely recognised standard in the field of computer vision for evaluating object detection models [27]. It quantifies the overlap between the predicted segmentation masks and the ground truth, providing a clear measure of the model's accuracy in delineating each object within the urban landscape (1). By computing the area of intersection divided by the area of union between the predicted and actual masks, IoU offers a precise and direct assessment of the model's performance. This metric is particularly effective for our application, as it caters to the varying shapes and sizes of urban elements, allowing for a nuanced evaluation of the segmentation results.

$$IoU = \frac{Area\ of\ Overlap\ between\ the\ Predicted\ and\ Actual\ Segmentation}{Area\ of\ Union\ between\ the\ Predicted\ and\ Actual\ Segmentation} \quad (1)$$

Upon applying the IoU metric to the fine-tuned OneFormer model's segmentation outputs, the results indicated a high level of accuracy in object detection (Table 2). The IoU scores consistently surpassed benchmark mean ones[1], affirming the model's capability to accurately identify and segment diverse classes of urban landscape features. These results underscore the effectiveness of the OneFormer model in understanding and processing complex urban scenes, which is critical for the application's intended use in urban landscape analysis and modification. The high IoU scores validate the model's technical proficiency and highlight its potential as a reliable tool for urban planners and designers, opening avenues for enhanced visualisation and creative reimagining of urban spaces.

| Class | Fine-tuned OneFormer | Benchmark mean |
|---|---|---|
| Buildings | 0.984 | 0.975 |
| Roads & pavements | 0.961 | 0.949 |
| Vehicles | 0.742 | 0.662 |
| Pedestrians & bicycles | 0.729 | 0.653 |
| Natural elements (trees, greenery, grass, etc) | 0.991 | 0.986 |
| Street furniture (poles, fences, streetlights, etc) | 0.807 | 0.695 |

Table 2: Validation results (IoU scores) for individual object detection

---

[1] For more details about benchmark scores, please see here: https://www.cityscapes-dataset.com/benchmarks/

## 5.2. Text input accuracy for image generation

The validation of image generation accuracy through text input is essential in determining the effectiveness of the fine-tuned SDXL model in accurately rendering diverse urban landscape elements, such as buildings, trees, bicycles, and more, based on textual descriptions. This process is vital to ensure that the generated images adhere to the specific details provided in the text prompts while maintaining a high degree of visual realism.

To comprehensively evaluate the image generation accuracy by text input, we employed the CLIP (Contrastive Language-Image Pretraining) methodology, as described in [28]. This method involves a dual assessment where both the generated images and a set of corresponding textual descriptions are encoded into a shared embedding space, enabling direct comparison of their similarity. The method consists of two main parts: an image encoder and a text encoder [28]. Each encoder transforms its respective inputs into a high-dimensional feature vector. The CLIP training objective is to maximise the cosine similarity for matching image-text pairs and minimise it for non-matching pairs. Mathematically, the cosine similarity between two vectors (u,v) is defined in (2):

$$Cosine\ Similarity(u,v) = \frac{u \cdot v}{\|u\|\|v\|} \qquad (2)$$

Post-training, when CLIP is used for evaluation, the method computes feature vectors for the input image and text using the pre-trained encoders, then calculates their cosine similarity. The resulting similarity score reflects how well the text describes the image according to the learned joint embedding space.

For our validation, we focused on a diverse range of urban object categories, namely: Buildings, Roads and pavements, Vehicles, Pedestrians and bicycles, Natural elements (e.g., trees, greenery, grass), and Street furniture (e.g., poles, fences, streetlights, traffic lights). For each category, we generated 40 images, totalling 240 images across all categories. Each of these images was then evaluated against a set of descriptive texts pertinent to their respective categories (e.g., "a glass building", "a busy road", "a group of tall trees").

In this process, the CLIP method computed similarity scores for each image-text pair. These scores, expressed as percentages ranging from 0 to 1, quantitatively reflect how well each generated image corresponds to the given textual descriptions. That is, higher CLIP scores imply higher correspondence; a CLIP score of 1 would mean that there is a perfect alignment between the image and the text description [28]. The evaluation was iterated across 8 cycles to observe trends and consistency in the model's performance. This methodical approach allowed us to gauge the precision of our fine-tuned SDXL model in aligning the generated visual content with the specified textual prompts, thereby assessing the model's efficacy in producing contextually accurate urban landscape imagery.

The validation results, as shown in Figure 2, reveal a clear upward trend in the accuracy of image generation across all urban object categories over the course of 8 iterations. Specifically, the category "Buildings" showed the highest scores, starting from an already impressive baseline and achieving near perfect alignment with the input text by the final iteration. "Pedestrians and bicycles" and "Vehicles" also demonstrated significant improvement, with scores steadily climbing, reflecting the model's increasing proficiency in these categories. In contrast, "Street furniture" started with the lowest scores but displayed consistent growth, indicating gradual learning and adaptation, albeit with a final score that still reflects room for improvement. Overall, the analysis of 40 images per category substantiated the model's capability to refine and adjust its output to closely mirror the textual descriptions, showcasing its utility in generating diverse and complex urban landscapes with high fidelity.

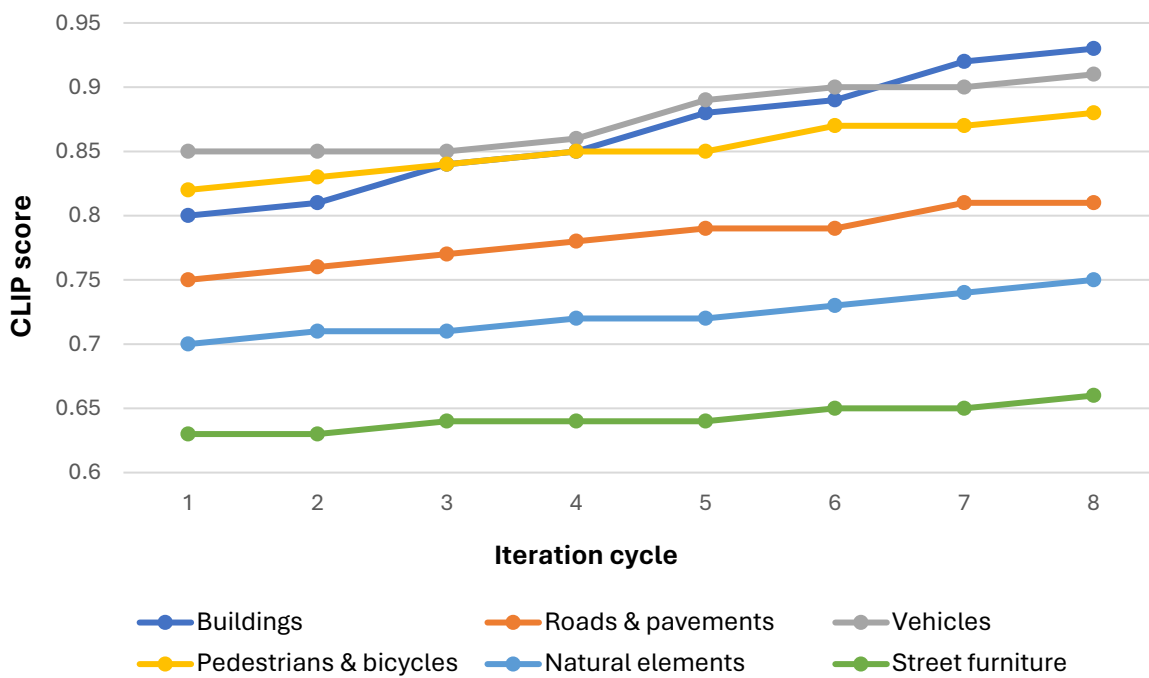

*Figure 2: Validation results (CLIP scores) for text-to-image generation accuracy*

## 6. Preliminary findings

The integration and testing of computational workflows within real-world contexts serve as a proof of concept and a means to gauge their practical utility and effectiveness. Therefore, we have applied our developed workflow in two distinct, yet equally significant, environments: as an educational aid in an academic setting and as a facilitative instrument for community engagement in urban design. These applications underscore the workflow's versatility and its potential in nurturing an inclusive, collaborative approach to urban landscape design.

## 6.1. UrbanGenAI as an educational tool

Within the Architectural Design Studio module at the University of Derby, students engaged with the prototype application as a centrepiece of a comprehensive workshop aimed at reimagining urban landscapes. This educational incorporation allowed students to apply theoretical knowledge to tangible design scenarios, using the application to visualise and iterate upon urban reconstructions based on their input images.

Figure 3 showcases a step-by-step application of the proposed workflow within an educational context, as utilised by students in an Architectural Design Studio module.

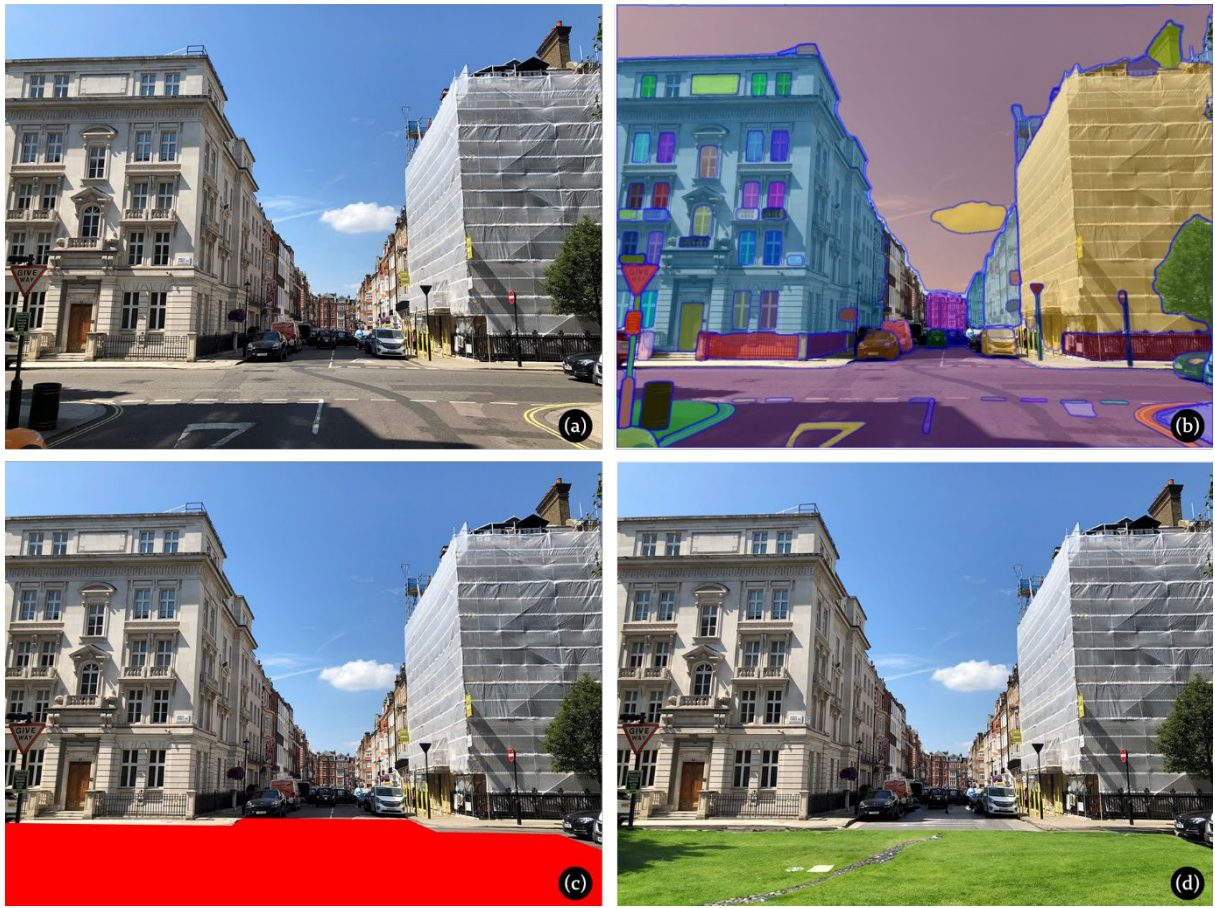

*Figure 3: Sequential workflow demonstration in architectural design education: (a) original urban landscape input by a student; (b) automated segmentation via OneFormer highlighting distinct urban elements; (c) focused object selection and masking; (d) final SDXL-generated image reflecting textual design prompt integration.*

Specifically, in the example presented in Figure 3:

> (a) The first panel displays the input image provided by a student, which captures a typical urban street scene. The image shows a mix of residential and commercial buildings, some of which are under construction and covered with scaffolding.

> (b) The second panel illustrates the segmented version of the input image, processed by the OneFormer model. This image is color-coded to differentiate between various elements such as buildings, the road, vehicles, and the sky. Each segment is outlined, demonstrating the model's ability to distinguish and separate the different components of the urban environment.
>
> (c) In the third panel, one particular object – i.e., the tarmac surface – is selected and masked. This portion of the image is highlighted in red, signifying the focus area for further processing.
>
> (d) The final panel presents the generated image after the application of the SDXL model in a ControlNet environment, based on the specified mask and accompanying textual prompt (i.e., "a grassy surface with cobbles"). The resulting image shows the tarmac surface having been digitally reconstructed, seamlessly blending back into the original street scene without the hard surface overlay.

The above example indicates the transformative capability of the workflow, from the initial real-world image through to the intelligent segmentation and culminating in a refined generated image that aligns with the design intentions of the student.

Preliminary findings suggest that the application fostered a deeper understanding of urban design dynamics, as students could immediately see the implications of their design decisions in a simulated real-world environment. The interactive nature of the tool also appeared to encourage experimentation and creativity, which are essential competencies in architectural education.

## 6.2. UrbanGenAI as a tool for co-designing with local communities

In a parallel vein, the prototype application was deployed as part of the Living Streets research study under the DUST research project, engaging the residents of Derby in the co-design of their neighbourhoods. This co-creational approach positioned the application as a bridge between the residents' lived experiences and the potential futures of their urban spaces.

Figure 4 is a visual representation of the workflow applied during a co-design session with local residents of Derby, forming part of the Living Streets study. Specifically, in the example presented in Figure 4:

> (a) The first panel shows the input image submitted by a resident, depicting a residential street in their neighbourhood. The scene captures houses, vehicles, and street elements under an overcast sky.
>
> (b) The second panel illustrates the output of the OneFormer segmentation, where various elements in the image have been identified and colour-coded. This segmentation delineates the different components of the street scene, such as the roads, pavements, vehicles, and natural elements.

(c) In the third panel, a specific object – i.e., a vehicle – has been selected by the resident and masked, highlighted in red against the rest of the neighbourhood scene.

(d) The final panel presents the image post-processing by the SDXL model in a ControlNet environment, where the selected object has been modified or potentially in response to the resident's textual input (i.e., "an array of flowerbeds"). The rest of the scene remains unchanged, maintaining the integrity of the neighbourhood environment.

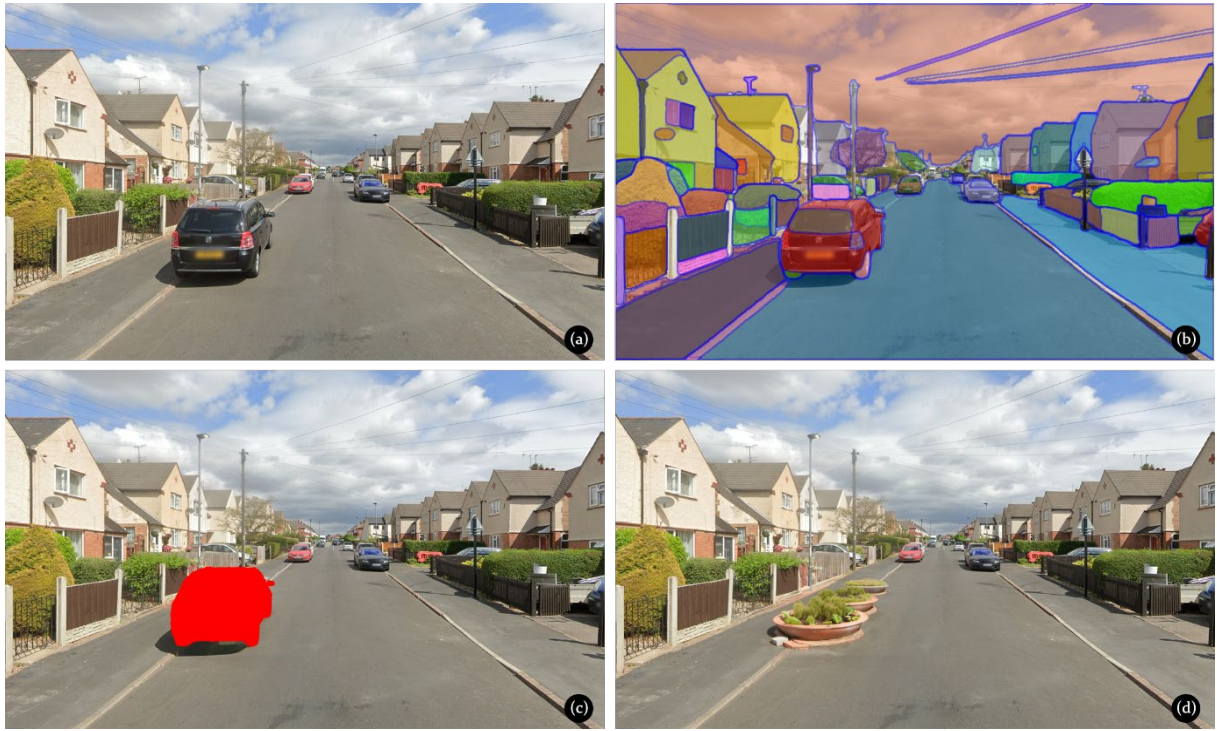

*Figure 4: Workflow application in community co-design: (a) original neighbourhood image contributed by a local resident; (b) detailed segmentation via OneFormer revealing distinct elements; (c) targeted object selection with an overlaid mask; (d) enhanced SDXL-generated image reflecting resident-specific modifications.*

This sequence effectively demonstrates the application's potential in engaging community members in the urban design process, allowing them to actively participate in the envisioning and reimagining of their living spaces.

The preliminary findings from this study highlight the tool's capacity to democratise the design process, empowering residents to contribute actively to the discourse on urban development. The visualisation capabilities provided by the application allowed for a shared understanding to emerge between residents and researchers, fostering a collaborative atmosphere conducive to insightful and meaningful urban landscape proposals.

## 7. Discussion

In this paper, we have outlined the development and application of UrbanGenAI, a novel workflow, encapsulated within a prototype desktop application, which harnesses computer vision and genAI capabilities for urban landscape reconstruction. The workflow integrates two pivotal components, i.e., the OneFormer model for detailed image segmentation of urban scenes and the SDXL diffusion model – fine-tuned and materialised through ControlNet – for image generation conditioned on masks and textual prompts. This comprehensive approach has undergone preliminary testing in two key areas: as an educational resource in the Architectural Design Studio module at the University of Derby, and as a participatory tool for co-creating urban design scenarios with local residents in the Living Streets research study. The deployment of UrbanGenAI within these contexts served as a preliminary testing ground to assess and confirm the effectiveness and versatility of both the proposed framework and the prototype application.

We have also evaluated the performance of the prototype application, in terms of object detection and text-to-image generation. Results indicated that UrbanGenAI exhibited a robust and promising capacity for urban landscape analysis and reconstruction. The fine-tuned OneFormer model demonstrated a keen ability to discern and segment a variety of urban elements. It proficiently identified and delineated buildings, roads, pavements, and natural elements within complex urban scenes. The precision in segmenting static objects like buildings and natural elements is matched by its adept handling of more transient and dynamic objects such as vehicles, and pedestrians & bicycles, showcasing the model's versatility.

Regarding text-to-image generation, the SDXL model's capacity was assessed through a series of evaluations, revealing its adeptness in translating textual prompts into corresponding visual outputs. The fine-tuned model showed a commendable ability to comprehend and visualise diverse urban elements as specified by various descriptive texts. This robustness in image generation, reflecting a wide spectrum of urban features, highlights UrbanGenAI's potential as a powerful tool for urban landscape design.

Early findings from the deployment of UrbanGenAI in educational settings have been indicative of its considerable potential as a learning tool in the pedagogy of architectural, urban, and landscape design. One of the most prominent qualities of our approach is its ability to enhance student creativity. By allowing students to visualise the immediate impact of their design choices, the tool encourages experimental and innovative thinking. It serves as a catalyst for imagination, enabling students to explore a plethora of design scenarios without the constraints of traditional design processes. Moreover, the application's highly interactive nature ensures that students are actively engaged in the learning process. They are not passive recipients of knowledge; rather, they become dynamic participants in a creative dialogue with the tool. This interactivity is further bolstered by the application's accessibility; it eliminates the barrier of complex software knowledge by simplifying the input to basic images and text prompts. This user-friendly interface democratises the design process, ensuring that students from various backgrounds can utilise the tool without the need for extensive training in specialised software.

Significantly, UrbanGenAI promotes environmental sustainability -- an increasingly critical consideration in contemporary design education. Trained on scenes depicting sustainable landscapes, the application inherently encourages students to consider environmentally responsible design solutions. This aspect of the tool aligns with the global shift towards sustainable practices, embedding these values in the educational phase. Finally, the application provides students with an integrated understanding of urban environments. It enables them to perceive the relationships and interdependencies between different urban features. Through the manipulation of these elements within the application, students can gain insights into the complex network of forces that shape urban spaces. This holistic sense of urban environments is essential for the training of future architects and urban designers, who will be tasked with crafting spaces that are not only aesthetically pleasing but also socially and ecologically responsive. The application, therefore, stands out as not just a technological advancement but as a potentially meaningful addition to design education as well.

Preliminary findings from the deployment of UrbanGenAI in community co-creation settings provided a compelling case for its use as a participatory planning and co-design tool. A standout quality of our approach is its facilitation of active participation in the design process. The application's intuitive interface invites input from individuals regardless of their technical expertise, lowering barriers to entry in the urban design discourse. This inclusivity is critical in participatory planning, as it ensures that the voices and visions of the actual residents, who are often laypersons in terms of design skills, are heard and visualised. UrbanGenAI serves as an easy-to-use platform for design communication and consultation, effectively translating residents' thoughts and preferences into visual representations. This visual communication is essential for consensus-building and shared understanding, allowing all stakeholders to see potential outcomes of planning decisions. It bridges the gap between professional urban designers and the broader community, fostering a collaborative environment where ideas can be discussed and refined collectively.

Furthermore, UrbanGenAI aligns with sustainability and well-being agendas, which are increasingly at the forefront of urban development. Having been trained on scenes that incorporate sustainable landscapes, the application naturally steers users towards environmentally conscious design choices. It acts as a subtle educator on the principles of sustainable urban living, prompting participants to consider how designs might promote ecological balance and community well-being. In essence, UrbanGenAI empowers community members to co-create their living spaces, imbuing the design process with democratic values and ensuring that the resultant urban landscapes are functional, aesthetically pleasing, and conducive to well-being.

As we look to the future, the roadmap for further developing and validating UrbanGenAI encompasses a multifaceted approach designed to maximise its utility and reach. To validate and refine our tool, an expanded series of case studies in both educational and co-design contexts are planned. These studies will enable us to assess UrbanGenAI's effectiveness across a variety of urban settings and user demographics, ensuring that it meets the diverse needs and expectations of its users.

In parallel with these studies, work is underway to transition our application to a web-based platform, enhancing its accessibility and collaborative potential. Two main features are under development for this new iteration. Firstly, a real-time feedback mechanism is envisioned, which will empower users to refine their inputs iteratively based on the outputs generated. This feature promises to lead to more precise and satisfying design outcomes. Secondly, we are exploring multi-user engagement features that will foster collective design efforts. By enabling peer input and feedback, these features will enrich the participatory nature of urban design and collaborative education, aligning with democratic ideals and inclusive practices.

Finally, we are considering the integration of 3D modelling capabilities and extended realities into our application. This development would significantly expand UrbanGenAI's functionality, providing users with a more comprehensive and immersive view of potential urban transformations. The addition of 3D modelling and extended realities is particularly pertinent for complex architectural, urban, and landscape designs, where understanding spatial relationships and the impact of design choices on the urban fabric is crucial. Together, these advancements will position UrbanGenAI at the forefront of urban design technology, offering a state-of-the-art resource for designers, educators, and communities alike.